\documentclass[lettersize,journal]{IEEEtran}
\usepackage{amsmath,amsfonts}
\usepackage{algorithmic}
\usepackage{algorithm}
\usepackage{array}
\usepackage[caption=false,font=normalsize,labelfont=sf,textfont=sf]{subfig}
\usepackage{textcomp}
\usepackage{stfloats}
\usepackage{url}
\usepackage{verbatim}
\usepackage{graphicx}
\usepackage{cite}
\usepackage{multirow}
\usepackage{hyperref}	

\begin{document}

\title{FoodSAM: Any Food Segmentation}
\author{Xing Lan, Jiayi Lyu, Hanyu Jiang, Kun Dong, Zehai Niu, Yi Zhang, Jian Xue

\textit{School of Engineering Science, University of Chinese Academy of Sciences}

{\tt \small \{lanxing19, lyujiayi21, jianghanyu23, dongkun22, niuzehai18, zhangyi214\}@mails.ucas.ac.cn}

{\tt \small xuejian@ucas.ac.cn}

        
\thanks{
This work was supported by the National Natural Science Foundation of China (62032022, 61929104, 62027827, 61972375) and Scientific Research Program of Beijing Municipal Education Commission (KZ201911417048).
\emph{(Corresponding authors: Jian Xue.)}}}

\markboth{Journal of \LaTeX\ Class Files,~Vol.~14, No.~8, August~2021}%
{Shell \MakeLowercase{\textit{et al.}}: A Sample Article Using IEEEtran.cls for IEEE Journals}

\IEEEpubid{0000--0000/00\$00.00~\copyright~2021 IEEE}

\maketitle

\begin{abstract}
In this paper, we explore the zero-shot capability of the Segment Anything Model (SAM) for food image segmentation. 
To address the lack of class-specific information in SAM-generated masks, we propose a novel framework, called \emph{FoodSAM}. 
This innovative approach integrates the coarse semantic mask with SAM-generated masks to enhance semantic segmentation quality.
Besides, we recognize that the ingredients in food can be supposed as independent individuals, which motivated us to perform instance segmentation on food images.
Furthermore, FoodSAM extends its zero-shot capability to encompass panoptic segmentation by incorporating an object detector, which renders FoodSAM to effectively capture non-food object information.
Drawing inspiration from the recent success of promptable segmentation, we also extend FoodSAM to promptable segmentation, supporting various prompt variants.
Consequently, FoodSAM emerges as an all-encompassing solution capable of segmenting food items at multiple levels of granularity. 
Remarkably, this pioneering framework stands as the first-ever work to achieve instance, panoptic, and promptable segmentation on food images.
Extensive experiments demonstrate the feasibility and impressing performance of FoodSAM, validating SAM's potential as a prominent and influential tool within the domain of food image segmentation. 
\end{abstract}



\section{Introduction}

The landscape of natural language processing \cite{wolf2019huggingface, ruder2019transfer, qiu2020pre} has been revolutionized by the emergence of large language models \cite{kasneci2023chatgpt, wei2022emergent, wei2022chain} trained on vast web datasets.
Notably, these models showcase impressive zero-shot generalization capabilities, enabling them to transcend their original training domain and exhibit proficiency across a spectrum of tasks and data distributions.
When coming to the domain of computer vision, the recent unveiling of the Segment Anything Project (SAM) by the Meta AI \cite{kirillov2023segment} 
introduces a groundbreaking promptable segmentation task, which is designed to train a robust vision foundation model. 
This ambitious work represents a significant advancement towards achieving comprehensive cognitive recognition of all objects in the world. 
The SAM project aims to investigate interactive segmentation challenges while effectively accommodating real-world constraints.

SAM demonstrates significant performance on various segmentation benchmarks, 
showcasing its impressing zero-shot transfer capabilities on 23 diverse segmentation datasets \cite{kirillov2023segment}.
In this paper, we investigate the zero-shot capability of the SAM for the domain of food image segmentation, a pivotal task within the field of food computing\cite{10.1145/3600095, zhou2019application, zhu2021deep}.
However, the vanilla mask generated by SAM alone does not produce satisfactory results, primarily due to its inherent deficiency in capturing class-specific information within the generated masks.
Moreover, compared to semantic segmentation on general object images, food image segmentation is more challenging due to the large diversity in food appearances and the imbalanced distribution of ingredient categories \cite{wu2021large}.
Consequently, accurately distinguishing the category and attributes of food items via SAM becomes a challenging task.

To this end, we propose a novel zero-shot framework, named FoodSAM\footnote{https://github.com/jamesjg/FoodSAM}, to incorporate the original semantic mask with SAM-generated category-agnostic masks.
While the SAM demonstrates remarkable food image segmentation capabilities, it is impeded by the absence of class-specific information.
Conversely, the conventional segmentation method preserves category information, albeit with a trade-off in segmentation quality.
To improve semantic segmentation quality, we advocate for the fusion of the original segmentation output with SAM's generated masks. 
Ascertaining the mask's category through the identification of its predominant elements is a novel and effective approach to enhance the semantic segmentation process.

Moreover, since the ingredients in the food are randomly cut and placed, they are supposed as independent individuals, which gives us the motivation to implement instance segmentation on food images. 
Therefore, the masks produced by SAM are inherently associated with individual instances, forming the foundation upon which we execute instance segmentation for food images.

Additionally, it is noteworthy that food images contain various non-food objects, such as forks, spoons, glasses, and dining tables.
Those objects are not the ingredients of the food, but they are also important for food, which reflects the attributes of the food.
To accomplish this purpose, FoodSAM introduces object detection methodologies \cite{padilla2020survey, zou2023object,zhou2022simple} to detect the non-food objects in the background.
By combining the no-used background masks generated by SAM, FoodSAM brings the object category labels for those masks as semantic labels.
In this manner, when coupled with our established instance segmentation methodology, the proposed framework enables the successful achievement of panoptic segmentation on food images.

Drawing inspiration from the SAM project, a noteworthy addition to our study involves the "prompt to food image segmentation", a fresh task proposed by SAM that augments the breadth of our investigation.
We design a simple but effective method for naive object detection to make FoodSAM support promptable segmentation.
In object detection, we convert the prompt learning method to prompt-prior selection and extend support to diverse prompt variants, such as point, box, and mask prompts.
We select interest objects according to whether the point is located, the box is covered, or the mask is overlapped.
At last, combined with SAM promptable segmentation and the original semantic mask, we also achieve promptable segmentation across multiple levels of granularity on both food and non-food objects.

\begin{figure*}[tbh]
\centering
\newpage
\includegraphics[width=\linewidth]{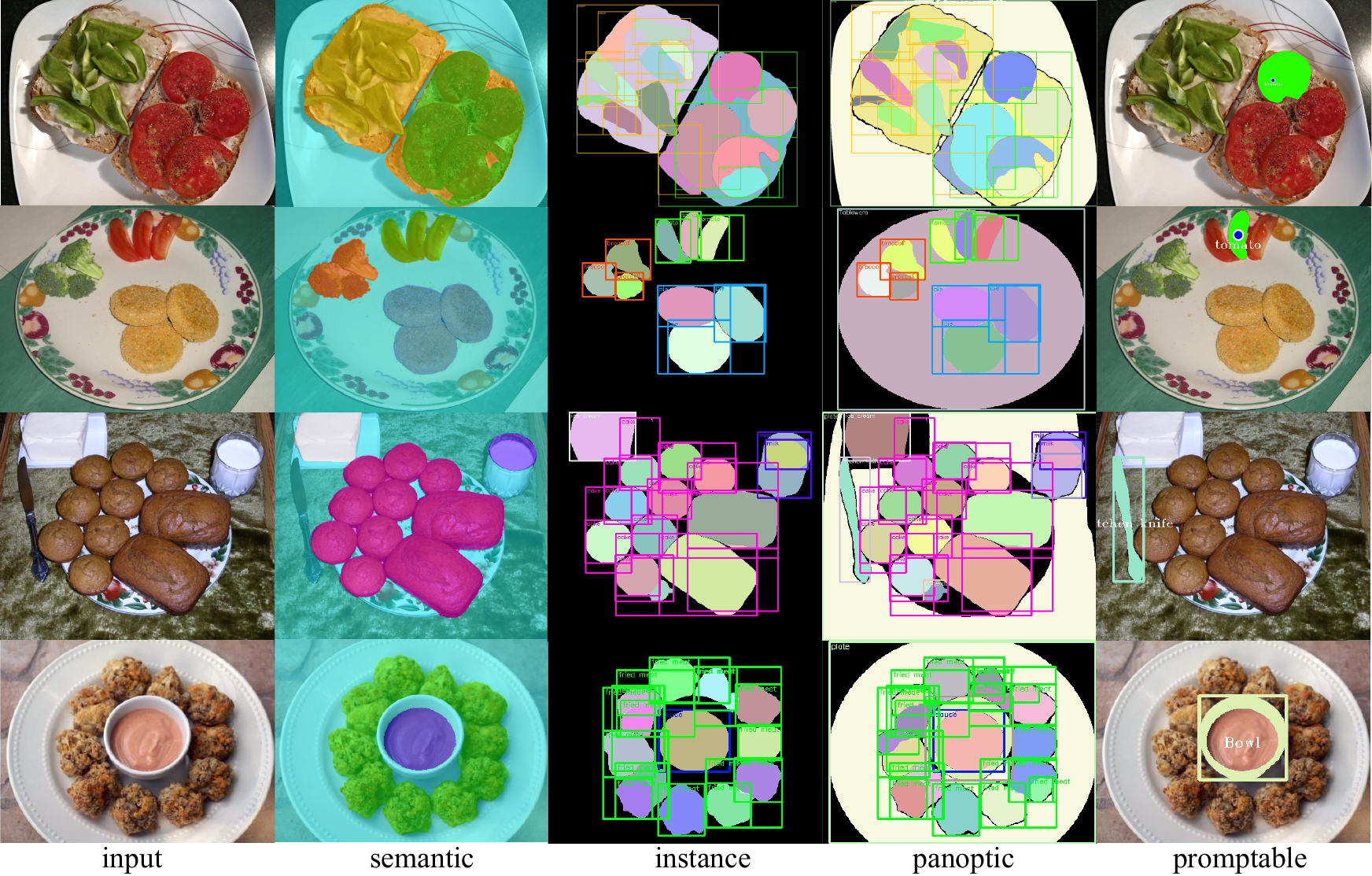}
\caption{FoodSAM emerges as an all-encompassing solution capable of segmenting food items at multiple levels of granularity. The different segmentation visualization is shown from left to right: input image, semantic, instance, panoptic and promptable, respectively. 
 }
\label{all seg vis}
\end{figure*}

Consequently, we present an all-encompassing method that can achieve any segmentation for food as shown in Fig.\ref{all seg vis}. 
In the literature on food image segmentation, this is the first work to accomplish instance segmentation, panoptic segmentation, and promptable segmentation.
Through a comprehensive evaluation on the FoodSeg103 \cite{wu2021large} and UECFoodPix Complete \cite{okamoto2021uec} benchmarks, 
FoodSAM outperforms the state-of-the-art methods on both datasets, highlighting the exceptional potential of SAM as an influential tool for food image segmentation.
In instance segmentation, our method achieves high-quality segmentation of individual ingredients. 
And in panoptic segmentation, non-food objects are also well-segmented with semantic labels.

The contributions of the paper are summarized as follows:
\begin{itemize}
    \item We present a novel zero-shot framework, named FoodSAM, which exhibits the unique capacity to perform food segmentation across diverse granularity levels.
    Significantly, our work represents the first exploration of applying the SAM to the domain of food image segmentation, effectively extending its zero-shot capabilities.
    
    \item To the best of our knowledge, this study stands as the first work to accomplish instance segmentation, panoptic segmentation, and promptable segmentation on food images.
    
    \item Experiments demonstrate the feasibility of our FoodSAM, which outperforms the state-of-the-art methods on both the FoodSeg103 and UECFoodPix Complete datasets. 
    Furthermore, FoodSAM performs better than other SAM variants on any food segmentation.
\end{itemize}


\section{Related Work}

\subsection{Foundation Model}
Foundation models, trained on broad data for adapting to diverse downstream tasks, have driven recent advances in machine learning. 
This paradigm often incorporates techniques like self-supervised learning, transfer learning, and prompt tuning. 
Natural language processing has particularly benefited, 
with the Generative Pre-trained Transformers \cite{bubeck2023sparks,floridi2020gpt} series pre-trained on massive text corpora enabling models \cite{brown2020language, hu2020gpt} to tackle translation, 
question answering, and other applications. 
Contrastive Language-Image Pre-training \cite{radford2021learning, li2021supervision} trained on image-text pairs 
can effectively retrieve images given text prompts, 
enabling image classification and generation applications. 
While most extract knowledge from available data, 
the Segment Anything Model \cite{kirillov2023segment} uniquely co-develops alongside model-in-the-loop annotation to construct a custom data engine 
with over 1 billion masks, 
conferring strong generalization. 
These foundation models \cite{bommasani2021opportunities, yuan2021florence, touvron2023llama} have achieved state-of-the-art performance across domains. 
The paradigm shows immense promise to profoundly advance machine learning across disciplines.

\subsection{Segmentation Task}
Image segmentation \cite{pal1993review,cheng2001color} encompasses various techniques to delineate distinct regions based on specified criteria. 
Each approach possesses unique characteristics and applications. 
Interactive segmentation leverages user guidance to enhance accuracy \cite{mcguinness2010comparative, mortensen1998interactive}
with users providing foreground/background markers to steer the algorithm. 
Superpixel methods group pixels into larger units called superpixels based on shared attributes like color and texture \cite{liu2011entropy}. 
This simplifies the image representation while retaining key structural aspects. 
Additional prevalent methods include thresholding, edge-based, region-based, 
and graph-based segmentation, exploiting intensity, discontinuity, similarity, and connectivity cues respectively \cite{haralick1985image}. 
The optimal technique depends on factors like task goals, data properties, and required outputs. 
Segmentation \cite{minaee2021image} remains an active area of research with existing methods limited in fully handling real-world complexity. 

Semantic segmentation represents a comprehensive approach in which each pixel is classified into a particular category, 
effectively partitioning the image according to semantic entities \cite{guo2018review, wang2018understanding, hao2020brief}. 
Recent methods like DANet \cite{fu2019dual}, OCNet \cite{yuan2018ocnet}, and SETR \cite{zheng2021rethinking} focus on extracting richer feature representations. 
Building upon semantic segmentation, instance segmentation \cite{hafiz2020survey} also delineates individual objects of the same class as distinct instances \cite{he2017mask,bolya2019yolact}. 
Panoptic segmentation unifies semantic and instance segmentation, 
assigning each pixel both a class label and a unique instance identifier if part of a segmented object \cite{kirillov2019panoptic, xiong2019upsnet, elharrouss2021panoptic}. 
This provides a holistic scene understanding by categorizing and differentiating all present elements. 
Each formulation provides unique information - semantic conveys categorical regions, instance delineates object quantities, 
and panoptic characterizes both detailed semantics and individual entity counts. 

Promptable segmentation has emerged as a versatile paradigm leveraging recent advancements in natural language models \cite{liu2023pre, luddecke2022image}. 
This approach employs careful "prompt engineering" to guide the model toward desired outputs \cite{liu2021gpt, lu2021pretrained}, 
departing from traditional multi-task frameworks with predefined tasks. 
At inference, promptable models adapt to new tasks using natural language prompts as context \cite{liu2023pre}. 
The Segment Anything Model \cite{kirillov2023segment} exemplifies this, 
training on 11 million images with 1.1 billion masks to segment objects specified in user prompts without additional fine-tuning. 
While SAM shows promising zero-shot generalization, a key limitation is the lack of semantic meaning in its mask predictions. 

\subsection{Food Segmentation}

Food image segmentation constitutes an essential and indispensable technology for enabling health applications such as estimating recipes, nutrition\cite{MIN2022100484}, and caloric content\cite{min2023large}. 
Compared to semantic segmentation of general objects, food image segmentation poses greater challenges due to the immense diversity in food appearances \cite{Min-ISIA-500-MM2020} and often imbalanced distribution \cite{klotz2021fine} of ingredient categories. 
For images containing multiple foods, segmentation represents a necessary precursor for dietary assessment systems. 
Segmenting overlapping, amorphous, or low-contrast foods lacking distinct color or texture features proves highly challenging. 
Furthermore, lighting conditions introducing shadows and reflections can adversely impact segmentation performance. 
Overall, the complex visual properties and arrangements of foods render this domain exceptionally demanding. 
Advanced segmentation techniques capable of handling food variability in unconstrained environments remain imperative for practical deployment.

Wu et al. \cite{wu2021large} proposed ReLeM to reduce the high intra-class variance of ingredients stemming from diverse cooking methods by integrating it into semantic segmentation models \cite{zheng2021rethinking,huang2019ccnet}.
Wang et al. \cite{wang2022swin} combined a Swin Transformer and PPM module in their STPPN model to achieve state-of-the-art food segmentation performance. 
Honbu et al. \cite{honbu2022unseen} explored zero-shot and few-shot segmentation techniques in USFoodSeg for unseen food categories. 
Sinha et al. \cite{sinha2023transferring} benchmarked Transformer and convolutional backbones for transferring visual knowledge to food segmentation. 
While these approaches have driven progress, semantic segmentation provides only categorical predictions without differentiating individual food items. 

\begin{figure*}[tbh]
\centering
\newpage
\includegraphics[width=\linewidth]{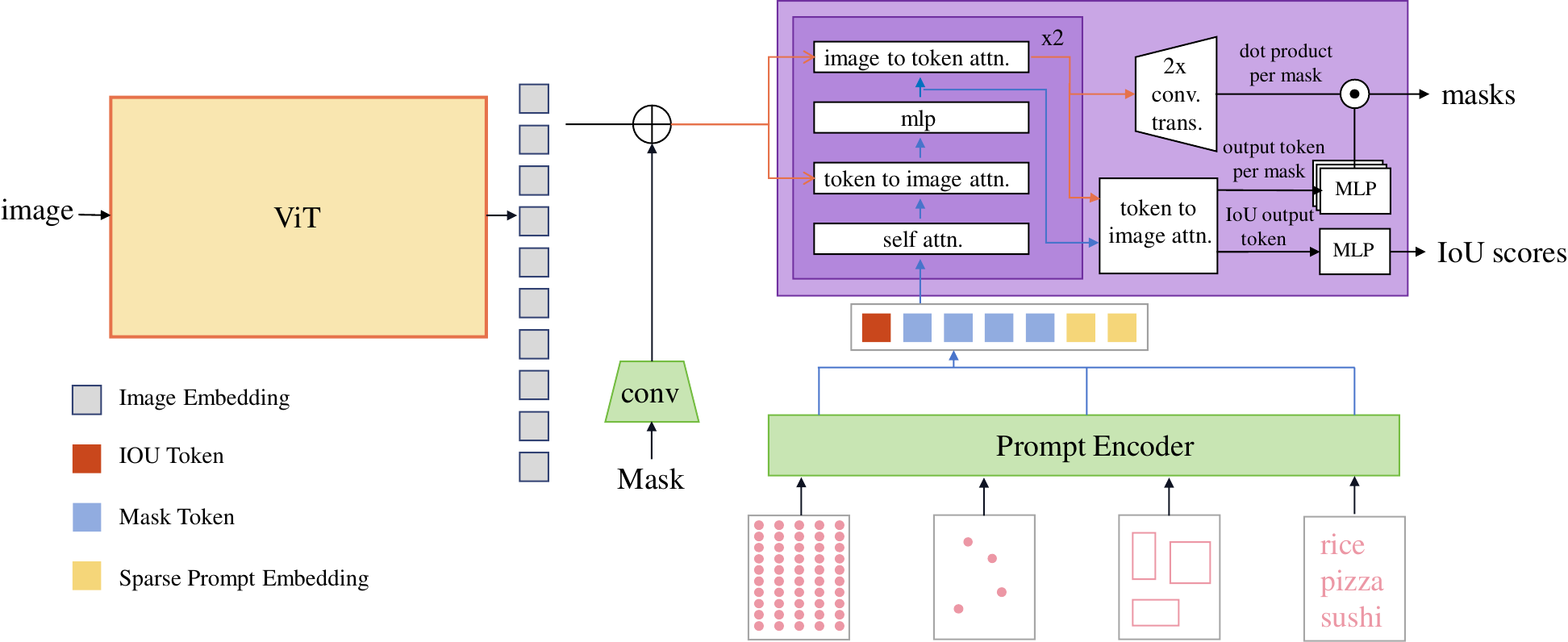}
\caption{The overview of Segment Anything Model (SAM) \cite{kirillov2023segment}. SAM contains three components: image encoder, prompt encoder, and mask decoder. }
\label{SAM pipeline}
\end{figure*}

However, to the best of our knowledge, no prior work has explored promptable or instance segmentation for food images and 
no existing datasets have supported instance-level or panoptic segmentation for food images. 
Semantic segmentation only provides categorical predictions without differentiating individual food items. 
In contrast, instance segmentation can delineate and count distinct food objects, enabling more accurate nutrition and calorie estimation. 
Furthermore, panoptic segmentation can characterize the surrounding environment, discerning attributes like the food's container and utensils. 
Such contextual cues provide meaningful signals about food properties and consumption habits. 
Therefore, advancing instance and panoptic food segmentation represent an important direction, 
as these task formulations are more informative than semantic alone for downstream food computing applications.

\section{Methodology}
\subsection{Preliminary}

\textbf{Revisit of SAM}: 
The Segment Anything Model (SAM) \cite{kirillov2023segment} represents the first application of foundation models to the image segmentation task domain. 
As shown in Fig.\ref{SAM pipeline} The proposed model is comprised of three key components - an image encoder, a prompt encoder, and a lightweight mask decoder module. 
The image encoder implements a computationally intensive vision transformer architecture with millions of parameters to effectively extract salient visual features from the input image. 
SAM provides three scale-specific pre-trained image encoder configurations: ViT-B (91M parameters), ViT-L (308M parameters), and ViT-H (636M parameters) \cite{dosovitskiy2020image,he2022masked}. 
The prompt encoder enables four types of textual or spatial inputs: points, boxes, freeform text, and existing masks. 
Points and boxes are represented using positional encodings\cite{vaswani2017attention}, and the text is encoded using a pre-trained text encoder from the CLIP model \cite{radford2021learning}, while mask inputs are embedded using convolutions. 
These prompt embeddings are summed element-wise with the image features. 
The mask decoder module employs a Transformer-based architecture, applying self-attention to the prompt and cross-attention between the prompt and image encoder outputs. 
This is followed by a dynamic mask prediction head that outputs pixel-level mask probabilities and predicted Intersection over Union (IoU) metrics. 
Transposed convolutions upsample the mask decoder features. 
Critically, the mask decoder can generate multiple mask outputs to account for inherent ambiguity in the prompt. 
The default configuration predicts 3 masks per prompt input. 
Notably, the image encoder extracts the feature only once per input image, allowing the cached image embedding to be reused across different prompts for the same image. 
This separation of the expensive image inference from the lightweight prompt interaction enables novel interactive use cases like real-time mobile Augmented Reality prompting. 
SAM was trained on a large-scale dataset of over 11 million images with 1 billion masks, yielding strong zero-shot transfer capabilities. 
As the name suggests, SAM can segment virtually any concept, even completely novel objects unseen during training.

\begin{figure*}[tbh]
\centering
\newpage
\includegraphics[width=\linewidth]{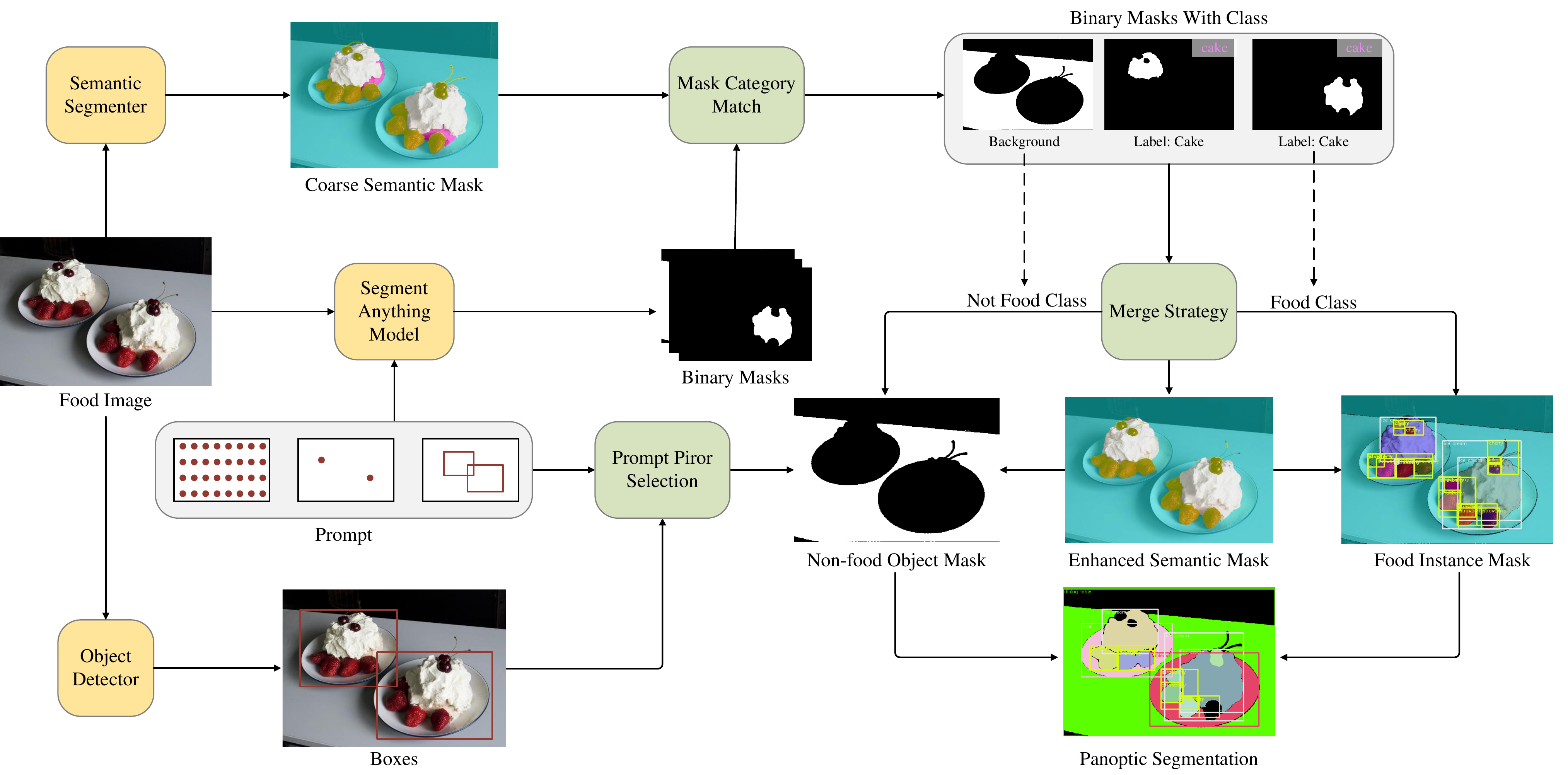}
\caption{The overview of our framework. FoodSAM contains three basic models: SAM, semantic segmenter, and object detector. SAM generates many class-agnostic binary masks, the semantic segmenter provides food category labels via mask-category match, and the object detector provides the non-food class for background masks. It then enhances the semantic mask via merge strategy and produces instance and panoptic results. Moreover, a seamless prompt-prior selection is integrated into the object detector to achieve promptable segmentation.}
\label{fig: foodsam}
\end{figure*}

\textbf{Variants of SAM}: 
Very recently, there are serval concurrent works proposed to address the limitations of SAM. Here we introduce them by following.

RAM \cite{zhang2023recognize} is an innovative image tagging foundational model based on SAM. Trained on a large set of image-text pairs, this model efficiently recognizes common categories, enabling the acquisition of a vast amount of image tags without the need for manual annotations.

SEEM \cite{zou2023segment} is an interactive segmentation model that can perform image segmentation in an all-pixel, all-semantics manner simultaneously while supporting interactive segmentation with various prompt types(including click, box, polygon, scribble, text, and referring region from another image). Their experiments demonstrate that SEEM achieves remarkable performance in open-vocabulary segmentation and interactive segmentation tasks, and exhibits robust generalization to diverse user intents.

SSA \cite{chen2023semantic} is a novel open framework that pioneers the application of SAM in the domain of semantic segmentation tasks. Its primary aim is to facilitate the seamless integration of users' pre-existing semantic segmenters with SAM, obviating the necessity for retraining or fine-tuning SAM's parameters. This integration empowers users to achieve enhanced generalization capabilities and finer delineation of mask boundaries in semantic segmentation tasks.

\subsection{Overview}
We explore applying SAM, a powerful mask generator, to food image segmentation. 
Although SAM segments food images with high quality, the generated masks lack categorical semantics. 
Standard semantic segmentation provides category labels but with low quality in food images. 

To address this, we proposed FoodSAM, a framework that merges the benefits of both approaches. 
We assign SAM's high-quality masks with semantic labels based on the mask-category match. 
Furthermore, due to the ingredients of food being randomly cut and placed when cooking, they are supposed as independent individuals,
which motivate us to achieve instance segmentation for food images.
Besides, to segment fine-grained objects apart from the background, FoodSAM contains an object detector for detecting non-food objects, such as a table, plates, spoons, etc. 
Therefore, FoodSAM also can achieve high-quality panoptic segmentation for food images.

Inspired by prompt learning, we also introduce promptable segmentation to food images, which is a novel task proposed by SAM. 
In the object detector, we convert the SAM's prompt learning way to the prompt-prior selection, which also supports the point, box, and mask prompts. 
In this way, FoodSAM supports interactive prompt-based segmentation across multiple levels of granularity on food images.

\subsection{FoodSAM}
FoodSAM consists of three main models: the segment anything model $M_a$, a semantic segmentation module $M_s$, and an object detector $M_d$, 
which also designs novel and seamless integration methods including the mask-category match, merge strategy, and prompt-prior selection. 
As shown in Fig.\ref{fig: foodsam}, we here describe them by the order of the pipeline: enhance segmentation, semantic-to-instance segmentation, and 
instance-to-panoptic segmentation. 
And promptable segmentation is inserted in this pipeline with its corresponding prompt.

\textbf{Enhance Semantic Segmentation}: Suppose the input food image $I\in R^{H\times W}$ forward by the semantic segmentation module model $M_s$ and output the semantic mask $m_s = M_s(I)$, 
which has the same shape as $I$ and the pixel value is corresponding to the category of the pixel.
SAM $M_a$ forward the image $I$ and output the binary candidate masks $m_a = M_a(I), m_a\in R^{K\times H \times W}$, where $K$ is the number of masks and the pixel value is True or False to indicate the pixel is foreground or background.

In the mask-category match, we use the i-th binary mask $m_a^i$ as the indices to get the i-th local semantic value set $ D_i =  m_s[m_a^i]$. 
Then, we use the voting scheme to choose the highest frequency semantic value $s_i$ in $D_i$ as the semantic label of the i-th mask $m_a^i$.
Meanwhile, we also calculate the confused degree of the category mask by
$d_i = \text{Count}(S_i, D_i) / \text{Len}(D_i)$, where $d_i$ denotes the i-th confused degree of category mask whose semantic label is $s_i$. 
And Count function summarizes the number of that label $s_i$ in the set $D_i$, Len function summarizes the number of the positive values $D_i$
 in the mask.  We also will filter the mask whose $d_i$ is lower than a threshold $\tau$, which means the mask label is confused and unstable.
 
Therefore, with the merge strategy, we can get the enhanced semantic mask $m_s^e$ by merging the $K$ binary masks $m_a^i$ with the semantic label $s_i$.
In this strategy,  we take consideration into the conflict area when those masks exist overlap. 
We first sort them by the area size $\text{Len}(D_i)$ and put them on the original semantic mask from large to small, which can preserve fine-object results
This process fully takes advantage of the high-quality mask of SAM and the semantic label of the semantic segmenter.

\textbf{Semantic-to-Instance Segmentation}: 
In the enhanced process, we get the semantic labels for each binary mask. 
Since the food is randomly cut and placed when cooking, we suppose the ingredients of food as independent individuals. 
Here we merge small masks into the nearby mask which has the same category label, and filter very small masks if their location is separate. 
After filtering the binary masks which correspond to the background category, we obtain the foreground masks. 
For those foreground semantic labels, the t-th binary mask $m_a^t$ with the voted semantic label $s_t$ is supposed as the t-th instance's mask $m_i^t$, where $t$ is the index of the foreground mask (instance id). 
Finally, we get the instance mask $m_i$ by merging the $T$ masks $m_i^t$ as well as the instance id.

\textbf{Instance-to-Panoptic Segmentation}: In the food image, the non-food objects, such as a table, plate, spoon, etc, are also important for the food image, 
which indicates the attributes of the food. 
Here we also suppose the non-food objects as independent individuals by introducing the object detector $M_d$.
The object detector $M_d$ forward the image $I$ and outputs the non-food bounding boxes $B_d$ with the corresponding class labels $C_d$.
In whole mask segmentation, SAM uses the dense gride points as the prompt, and remain the $K$ points which corresponding to $m_a$.
Here, we take each point of SAM-generated mask to select the non-food object candidates by judging if located in the bounding box $B_d$.
We next collect the binary masks whose semantic label is background, and then calculate the IoUs of the non-food candidate bounding boxes and that binary mask. 
When the highest IoU is large than another threshold, we suppose the corresponding class is the category of that mask. 
Otherwise, it will be preserved its background label. 
And non-food masks also will be merged followed by semantic-to-instance practice. 
Finally, we get the panoptic mask $m_p$ by merging the instance mask $m_i$ and the non-food object's semantic masks.

\textbf{Promptable Segmentation}: SAM proposes a novel task promptable segmentation, which can achieve interactive prompt-based segmentation. 
In the FoodSAM, we also introduce promptable segmentation to food images and further estimate the food category.
By leveraging the original semantic segmentation, we first segment interest food objects by SAM and can achieve interactive prompt-based segmentation at semantic and instance levels. 
Moreover, we propose a novel design for the object detector $M_d$ to support promptable detection across multiple levels of granularity. 
In the object detector, we convert the SAM's prompt learning way to prompt-prior selection for the interest object. 
The concurrent work FastSAM\cite{zhao2023fast} also introduces a selection strategy, but it is used after finished segmentation.

Different from FastSAM, we use the prompt-prior selection to segment the non-food objects in the background. 
It mainly involves the utilization of point prompt, box prompt, and mask prompt. 
Those non-food objects in the background are all assigned the category label from object detector.
For the point prompt, we first remain object's box information as prior boxes and use the point prompt to check the points whether in the prior boxes or not and most center to select the prior box. 
For the box prompt, we take IoU matching between the box prompt and the detected box from the object detector. 
The goal is to find the prior box with the highest IoU with the detected box. 
For the mask prompt, we sample points in the mask and check the points whether in the detected box, then followed by the same way as the point prompt. 
Specifically, for the regular prompt (the mask shape is the same as the input image), it uses near all non-food information which contain $K$ points for the next segmentation. 
Consequently, our FoodSAM achieves promptable segmentation on both food and non-food objects across multiple levels of granularity.



\section{Experiments}

\subsection{Experiment Settings}

\subsubsection*{Datasets}

    UECFoodPix Complete \cite{okamoto2021uec} was released in 2020 by the University of Electro-Communications.
    It includes 102 dishes and comprises 9000 training images as well as 1000 test images.
    And semantic labels are provided for each food item with 103 class labels.
    The segmentation masks were obtained semi-automatically using GrabCut, which segments images based on user-initialized seeding \cite{rother2004grabcut}. 
    The automatically-generated masks were further refined by human annotators based on a set of predefined rules \cite{ege2019new}.

    FoodSeg103 \cite{wu2021large} is a recent dataset designed for food image segmentation, consisting of 7,118 images depicting 730 dishes. 
    FoodSeg103 aims to annotate dishes at a more fine-grained level, capturing the characteristics of each dish’s individual ingredients.
    Specifically, the training set contains 4983 images with 29530 ingredient masks, while the testing set contains 2135 images with 12567 ingredient masks. 
    They were obtained through manual annotations. 
    Compared to UECFoodPixComplete, FoodSeg103 proves to be a more challenging benchmark for food image segmentation.
    Further, unlike UECFoodPixComplete, which covers entire dishes but lacks fine-grained annotation for individual dish components,

\subsubsection*{Implementation Details}
We have conducted experiments on the two datasets mentioned above with NVIDIA GeForce RTX 3090 GPU. 
For the components of FoodSAM, we use the ViT-h \cite{dosovitskiy2020image} as SAM's image encoder, and same hyperparameters as the original paper. 
For object detector, we use UniDet\cite{zhou2022simple} with Unified learned COIM RS200. 
For semantic segmentation, we use SETR\cite{zheng2021rethinking} as the baseline, which is with ViT-16/B as encoder, MLA as a decoder in FoodSeg103, and the checkpoint is provided by the GitHub repo. 
In UECFoodPix Complete, we take the deeplabv3+ \cite{chen2018encoder} as the baseline, and the checkpoint is retraining with the same hyperparameters as the paper reported.

\begin{figure*}[tbh]
\centering
\newpage
\includegraphics[width=\linewidth]{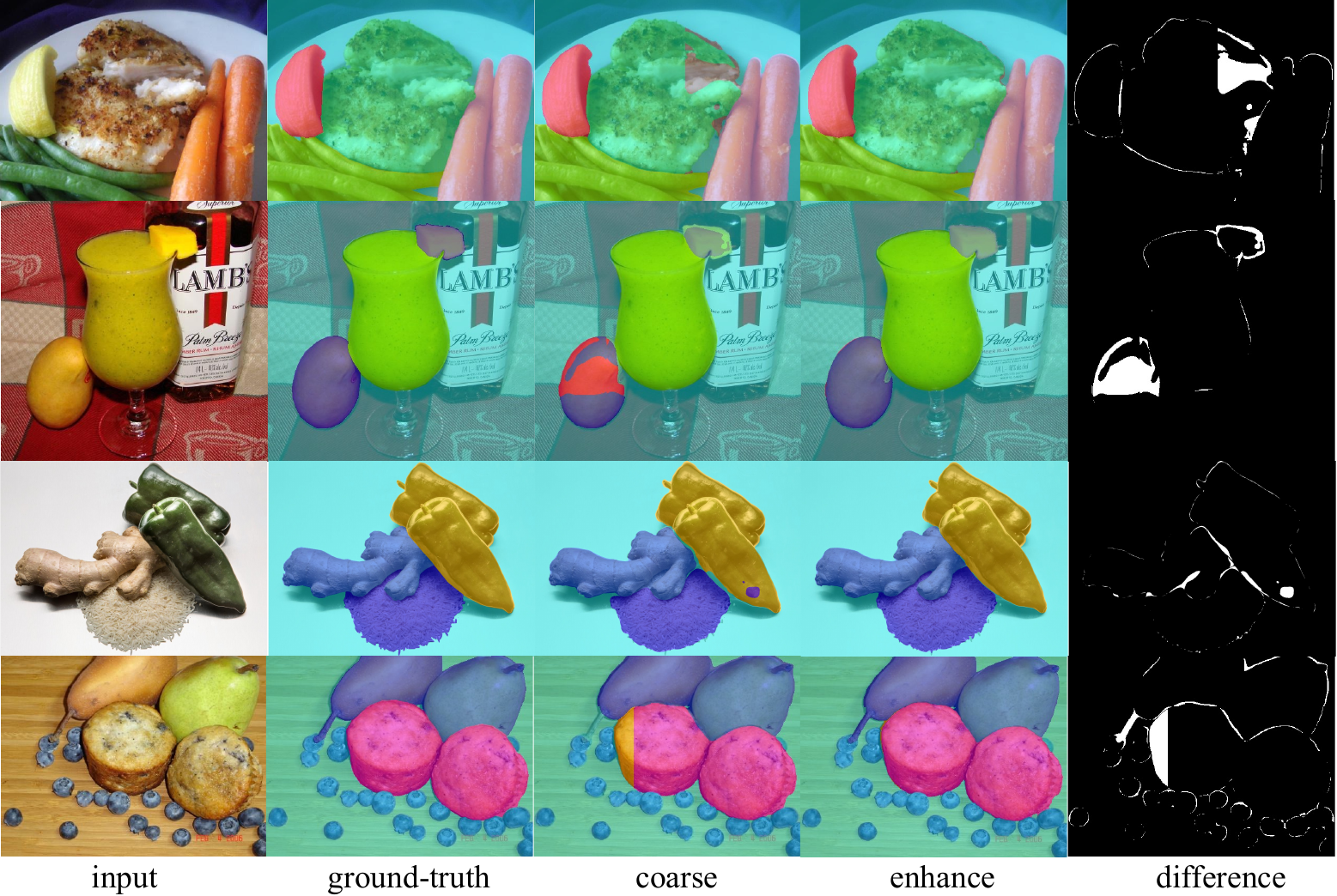}
\caption{Visualization comparison with baseline and ground-truth on semantic segmentation. The difference is calculated between the enhanced and coarse.}
\label{fig: semantic vis vs. baseline}
\end{figure*}

\subsubsection*{Evaluation Metrics}

We evaluate the performance of our model with some common metrics, e.g. mIoU(mean IoU over each class), mAcc(mean accuracy over all classes), and aAcc(over all pixels). mIoU is a standard indicator in semantic segmentation to assess the overlap and union between inference and ground truth, which is depicted by the following.
\begin{equation}
    \text{mIoU} = \frac{1}{N} \sum_{i=1}^N \frac{\mathrm{TP}_i}{\mathrm{TP}_i + \mathrm{FP}_i + \mathrm{FN}_i}
\end{equation}
where N is the number of classes, $\mathrm{TP}_i$, $\mathrm{FP}_i$, and $\mathrm{FN}_i$ are described as follows.
\\$\bullet$ True Positive ($\mathrm{TP}_i$) represents the number of pixels that are correctly classified as class $i$.
\\$\bullet$ False Positive ($\mathrm{FP}_i$) denotes the number of pixels that are wrongly classified as class $i$.
\\$\bullet$ False Negative ($\mathrm{FN}_i$) is the number of pixels that are wrongly classified as other classes while their true labels are class $i$. 
\\ mAcc is the average accuracy of all categories. For a dataset with N classes, it can be formulated as:
\begin{equation}
    \text{mAcc}=\frac{1}{N}\sum_{i=1}^N\frac{\mathrm{TP}_i}{\mathrm{TP}_i + \mathrm{FN}_i}
\end{equation}
\\And aAcc directly calculates the ratio of all pixels that are correctly classified, which can be described as:
\begin{equation}
    \text{aAcc} =\frac{\sum_{i=1}^N\mathrm{TP}_i}{\sum_{i=1}^N\mathrm{TP}_i + \mathrm{FN}_i}
\end{equation}

\begin{table}[]
\centering
\caption{Comparison with State-of-the-art Methods on FoodSeg103}
\begin{tabular}{c|ccc}
\hline
\multicolumn{1}{c}{\multirow{2}{*}{Methods}} & \multicolumn{3}{c}{Metrics}                   \\ \cline{2-4} 
\multicolumn{1}{c}{}                         & \textbf{mIoU (\%)} & \textbf{aAcc (\%)} & \textbf{mAcc (\%)} \\ \hline
FPN \cite{kirillov2019panoptic}              & 27.28         & 75.23         & 36.7          \\
CCNet \cite{huang2019ccnet}                   & 28.6          & 78.9          & 47.8          \\
ReLeM-CCNet \cite{wu2021large}                & 29.2          & 79.3          & 47.5          \\
ReLeM-FPN-Finetune \cite{wu2021large}                          & 30.8          & 78.9          & 40.7          \\
Window Attention \cite{dong2021windows}       & 31.4          & 77.62         & 40.3          \\
Upernet \cite{xiao2018unified}                & 39.8          & 82.02         & 52.37 \\
STPPN \cite{wang2022swin}                     & 40.3          & 82.13         & 53.98         \\
CCNet-Finetune                                & 41.3          & 87.7          & 53.8          \\
\hline
SETR \cite{zheng2021rethinking} \emph{(baseline)}   & 45.1          & 83.53         & 57.44         \\
\textbf{FoodSAM (ours)}                        & \textbf{46.42}            & \textbf{84.10}            & \textbf{58.27}             \\ \hline
\end{tabular}
\captionsetup{justification=centering}
\label{tab: foodseg103 sota}
\end{table}

\begin{table}[]
\centering
\caption{Comparison with State-of-the-art Methods on Uecfoodpix Complete}
\begin{tabular}{c|ccc}
\hline
\multicolumn{1}{c}{\multirow{2}{*}{Methods}} & \multicolumn{3}{c}{Metrics}                   \\ \cline{2-4} 
\multicolumn{1}{c}{}                         & \textbf{mIoU (\%)} & \textbf{aAcc (\%)} & \textbf{mAcc (\%)} \\ \hline
deeplabV3+ \cite{chen2018encoder,okamoto2021uec}             & 55.50 & 66.80 & --    \\
YOLACT \cite{battini2023segmented}                 & 54.85 & --    & --    \\
GourmetNet\cite{sharma2021gourmetnet}             & 62.88 & 87.07    & 75.87    \\
BayesianDeeplabv3+\cite{aguilar2022bayesian}     & 64.21 & 87.29 & 76.15 \\
\hline
deeplabV3+ \emph{(baseline)} & 65.61 & 88.20 & 77.56 \\
\textbf{FoodSAM (ours)} & \textbf{66.14}    & \textbf{88.47}    & \textbf{78.01}    \\ \hline
\end{tabular}
\captionsetup{justification=centering}
\label{tab:uec sota}
\end{table}

\subsection{Comparison with State-of-the-art Methods}
To compare with state-of-the-art methods, we have conducted experiments across multiple levels of granularity containing semantic, instance, and panoptic. 
Also, we compare our FoodSAM with SAMs in promptable segmentation tasks. 
We next discuss the results in detail.

\subsubsection*{Evaluation on Semantic Segmentation}

\begin{figure*}[tbh]
\centering
\newpage
\includegraphics[width=\linewidth]{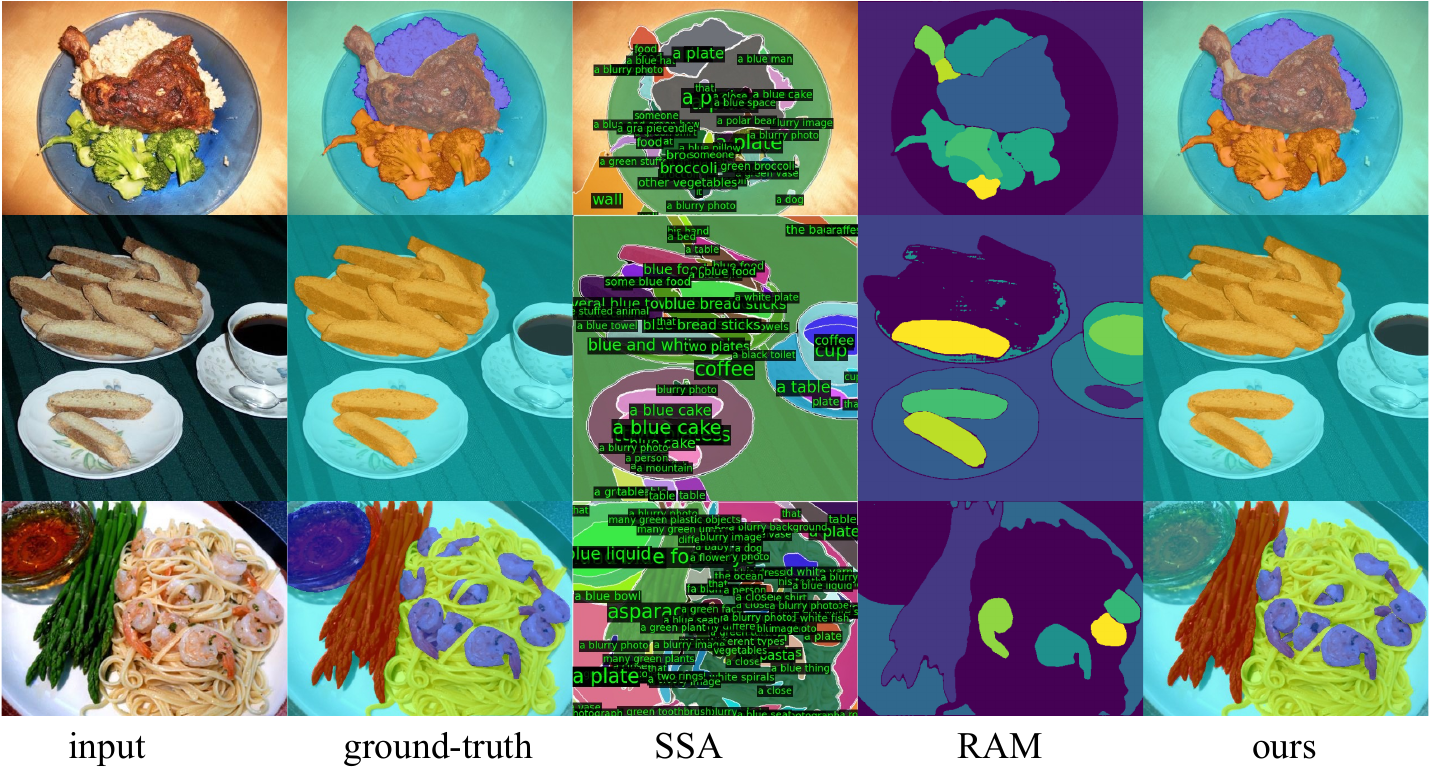}
\caption{Visualization comparison with SSA and RAM on semantic segmentation. SSA and RAM may output the instance information and their results are obtained by the public repository, here we only discuss semantic results on food.}
\label{fig: semantic vis vs. sams}
\end{figure*}

\begin{figure*}[tbh]
\centering
\newpage
\includegraphics[width=\linewidth]{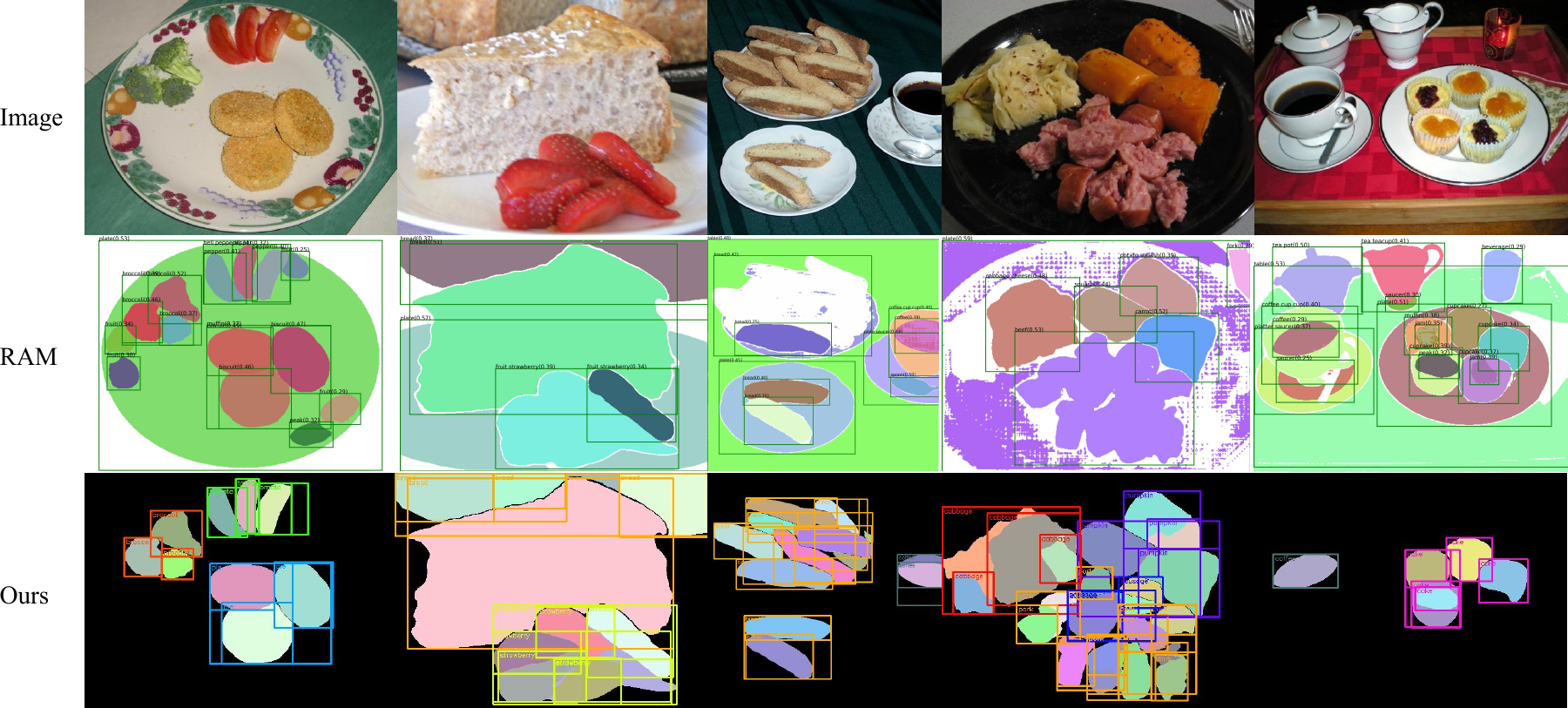}
\caption{Visualization comparison with RAM on instance segmentation. RAM may output the non-food instance information, here we only discuss the semantic results on food.}
\label{fig: instance vis vs. sams}
\end{figure*}

As shown in Tab. \ref{tab: foodseg103 sota}, we have achieved SOTA performance in FoodSeg103, 
As shown in Tab. \ref{tab:uec sota}, we have achieved SOTA performance in UECFoodPix Complete. 
Specifically, FoodSAM achieves 46.42 mIoU, 58.27 mAcc, 84.10 aAcc on FoodSeg103 as well as 66.14 mIoU, 78.01 mAcc and 88.47 aAcc on UECFoodPix.
And we also test the performance with other zero-shot methods as depicted in Tab. \ref{tab: foodseg103 zero shot}, FoodSAM outperforms recent SAM variants. 
Notably, the performance of these zero-shot methods all reach under 30 mIoU, which even performs the best supervision method at 45.1 mIoU.

As depicted in Fig.\ref{fig: semantic vis vs. baseline}, we also conduct the qualitative analysis for the original semantic mask, enhanced semantic mask, and ground-truth mask. 
By leveraging the impressive segment capacity of SAM, FoodSAM presents a powerful semantic segmentation method to compensate for the loss by the original segmenter. 
We also visualize the comparison with other SAM variants SSA and RAM as shown in Fig.\ref{fig: semantic vis vs. sams}, which shows that FoodSAM can segment finer ingredients and RAM has more mis-segmentation cases.

\begin{table*}[htbp]
\centering
\caption{Comparison with Zero-shot Methods on FoodSeg103}
\resizebox{\textwidth}{!}{
\begin{tabular}{c|c|ccccccccc|c}
\hline
\textbf{Methods}  & \textit{\textbf{Best sup.(UB)}} & ZSSeg-B & ZegFormer-B & X-Decoder-T & SAN-B & OpenSeeD-T   & OVSeg-L & SAN-L  & CAT-Seg-H & Gr.-SAM-H & \textbf{FoodSAM (ours)} \\ \hline

mIoU (\%) & 45.1 & 8.17 & 10.01 & 2.61 & 19.27 & 9.0  & 16.43 & 24.46  & 29.06 & 9.97 & \textbf{46.42} \\ 

\hline

\end{tabular}
}
\captionsetup{justification=centering}
\label{tab: foodseg103 zero shot}
\end{table*}

\subsubsection*{Evaluation on Instance Segmentation}

Due to that no previous work has implemented instance segmentation in that area, 
and no related benchmark has been released, we here compare the qualitative results with SAM variants. 
As shown in Fig. \ref{fig: instance vis vs. sams}, FoodSAM shows an impressive performance that can recognize well the instance identity of food with a pred semantic mask. 
And compared with other methods, FoodSAM can segment and recognize the special food ingredients that other works lack.
There exist some mis-segmentation cases in RAM \cite{zhang2023recognize}, e.g. the mis-segmentation of the fruit and bread.
And the strawberry is segmented as a whole not fine-grained enough in RAM, but FoodSAM can segment the strawberry into each small piece.

\subsubsection*{Evaluation on Panoptic Segmentation}

\begin{figure*}[tbh]
\centering
\newpage
\includegraphics[width=\linewidth]{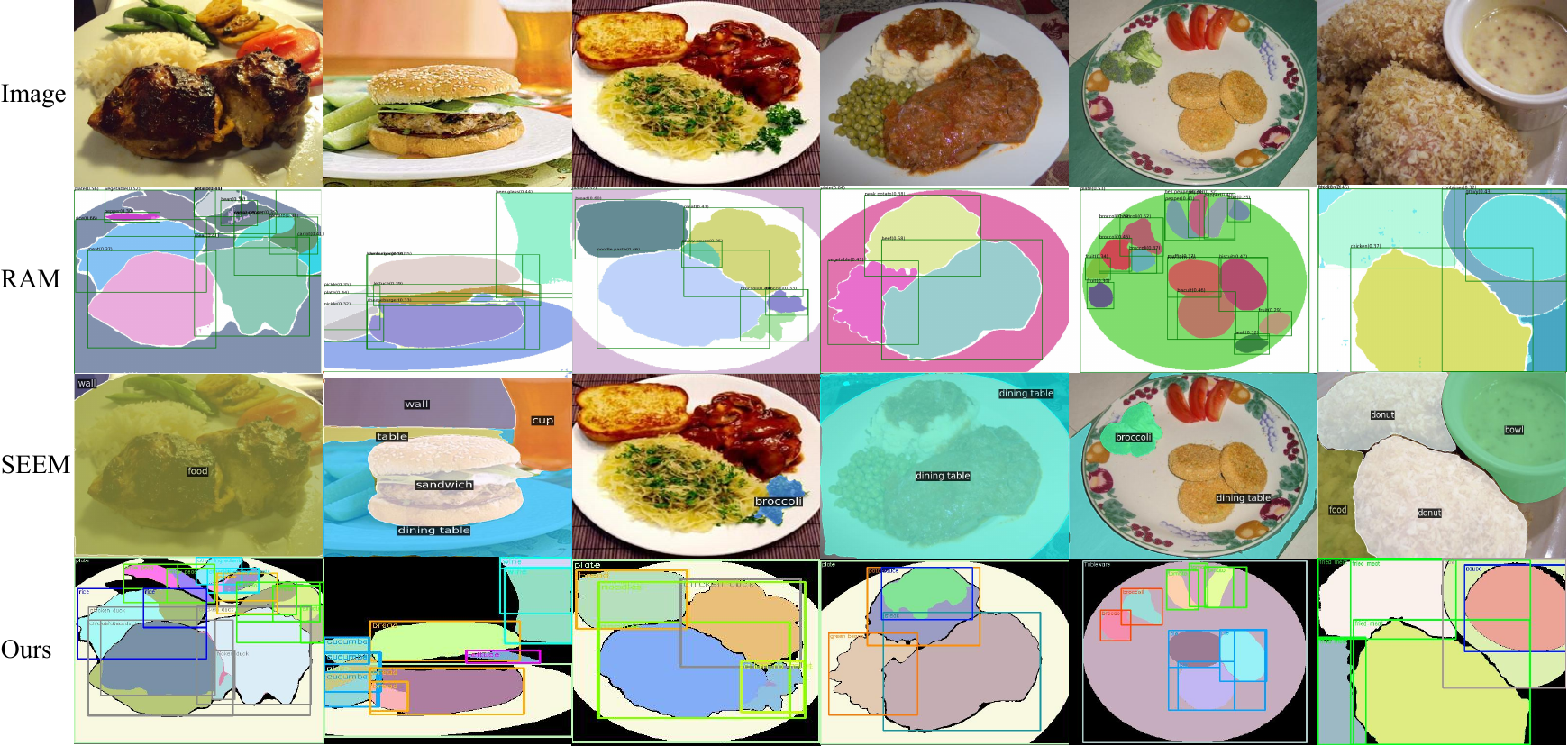}
\caption{Visualization comparison with RAM, SEEM and ours on panoptic segmentation. The visualization results are obtained from their public code repository.}
\label{fig: panoptic vis vs. sams}
\end{figure*}

Also, since no work and datasets have been public in this task, we conduct the qualitative analysis with SAM variants RAM and SEEM. 
As shown in Fig. \ref{fig: panoptic vis vs. sams}, FoodSAM performs an excellent segment ability for non-food objects. 
Specifically, for the same input image, it shows the additional bowl and plate compared with instance segmentation that focuses on the food. 
Compared with other methods, that cannot effectively distinguish the fine-grained difference such as a bowl with the ingredients and a glass with milk.

\subsubsection*{Evaluation on Promptable Segmentation}

\begin{figure*}[tbh]
\centering
\newpage
\includegraphics[width=\linewidth]{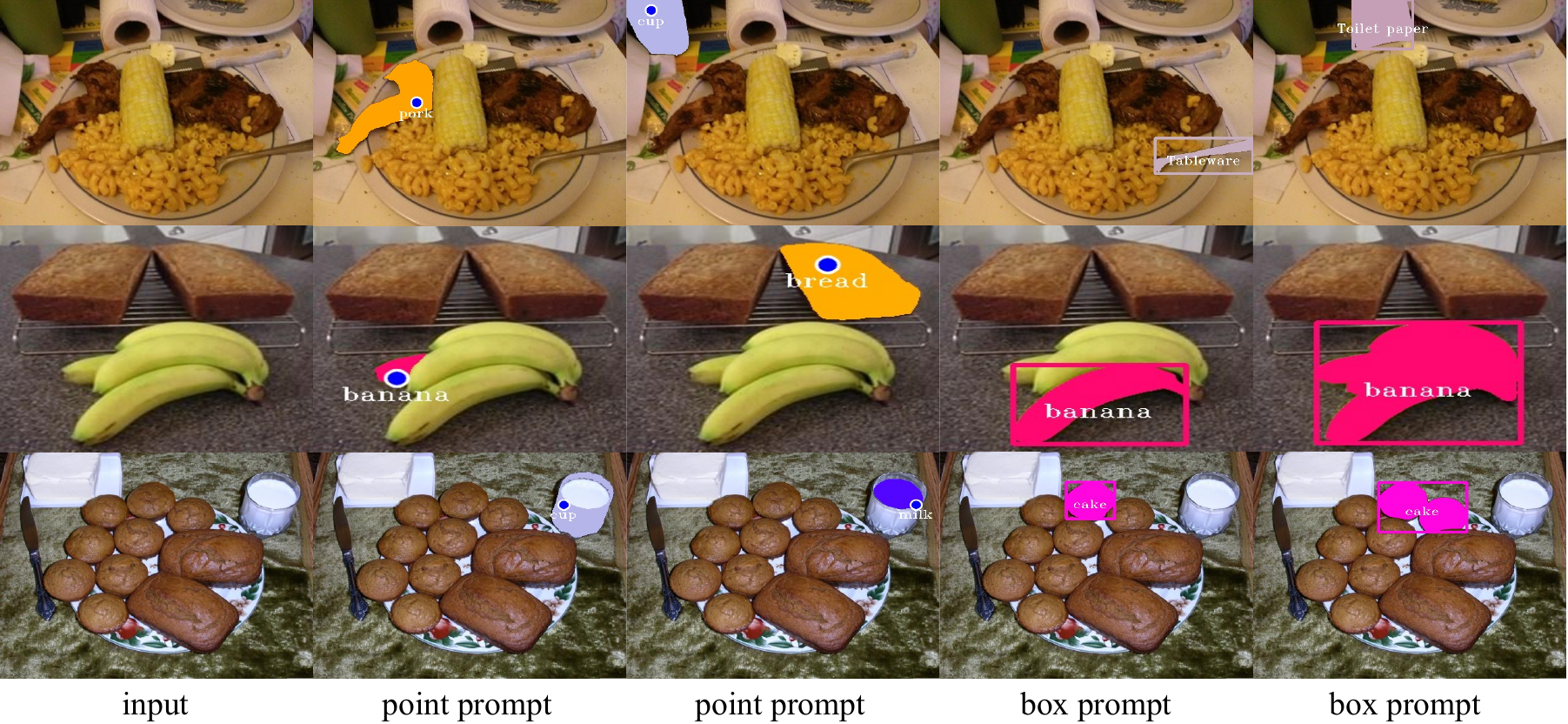}
\caption{Visualization results on promptable segmentation. From left to right: input, double point prompts, double box prompts}
\label{fig: prompt vis}
\end{figure*}

Drawing inspiration from SAM, we also extend the promptable segmentation to FoodSAM.
The regular or mask prompt is discussed in the above subsections. 
And here we discuss the point prompt and box prompt in this subsection, we direct use panoptic segmentation to compare granularity for it contains all fine-grained information, including food and non-food objects.
As shown in Fig.\ref{fig: prompt vis}, FoodSAM can identify the category of the food ingredients, which addresses the limitation of SAM. 
Moreover, FoodSAM also recognizes and segments the non-food object within the coarse background with an impressing result.

\subsection{Improvement of FoodSAM}
In this subsection, we also explore the improvement by leveraging the SAM segment performance. 
As shown in Tab.\ref{tab: foodseg103 sota} and Tab.\ref{tab:uec sota}, FoodSAM achieves obvious improvement compared with baselines, which indicates that SAM performs impressing segment performance and compensates for the original semantical segmenter limitation.
SAM produces many masks for each may exist object, we here experiment with the improvement of the number of merging masks.

Specifically, we choose the top-k area masks from FoodSAM to merge with the original semantic mask from the baseline. 
As shown in Tab.\ref{ablation: topk masks} and Tab.\ref{ablation: uec topk masks}, the performance increase with the increasing k.
When using 80 masks, FoodSAM achieves 66.14 mIoU, 78.00 mAcc, 88.47 aAcc on UECFoodPix Complete and 46.42 mIoU, 58.27 mAcc, 84.10 aAcc on FoodSeg103.
These experiments demonstrate that it is feasible for the fusion between SAM and food segmenter, which shows significant improvement with FoodSAM. 

\begin{table}
\centering
\caption{Results about merging original semantic mask and top K SAM masks on FoodSeg103}
\begin{tabular}{c|ccc}
\hline
\textbf{Top K}             & \textbf{mIoU (\%)} & \textbf{mAcc (\%)} & \textbf{aAcc (\%)} \\ \hline

Baseline   & 45.10         & 57.44         & 83.53         \\
+10 & 45.22         & 57.19         & 83.63         \\
+30 & 46.19         & 58.01         & 84.03 \\
+40 & 46.32         & 58.15         & 84.06         \\
+80 & 46.42         & 58.27         & 84.10         \\ \hline
\end{tabular}
\captionsetup{justification=centering}
\label{ablation: topk masks}
\end{table}

\begin{table}
\centering
\caption{Results about merging original semantic mask and top K SAM masks on Uecfoodpix Complete}
\begin{tabular}{c|ccc}
\hline
\textbf{Top K}             & \textbf{mIoU (\%)} & \textbf{mAcc (\%)} & \textbf{aAcc (\%)} \\ \hline

Baseline   & 65.61         & 77.56         & 88.2         \\
+10 & 66.01         & 77.89         & 88.39         \\
+30 & 66.12         & 77.99         & 88.46 \\
+40 & 66.13         & 78.00         & 88.46         \\
+80 & 66.14         & 78.00         & 88.47         \\ \hline
\end{tabular}
\captionsetup{justification=centering}
\label{ablation: uec topk masks}
\end{table}

\subsection{Ablation Study}

To verify the function of each part, we also implement the ablation study in these parts on FoodSeg103 benchmark: \textbf{f}iltering the mask with a \textbf{c}onfused \textbf{c}ategory label (FCC) or not, and \textbf{w}ith \textbf{S}AM-generated \textbf{m}ask (WSM)  or not. 
In only WSM, we use zero-mask as the init background mask and the label from the original semantic mask, while in the case of combining the baseline, we also use the original semantic mask as the init background. 
As shown in Tab.\ref{tab:ablation_all}, only WSM also could achieve comparable performance and the combination could get significant improvement compared to the baseline. With FCC, we merge with useful category label, and the mask with confused category use the corresponding area with semantic meanings.

And in the merging process, it needs to deal with the conflict area when masks exist the overlap. We also conduct an ablation study to verify which way is most effective, which considers two factors: SAM-predicted IoU and the number of positive values in a mask. 
As shown in Tab.\ref{tab:ablation_sortway}, sorted by area, from large to small, achieves the best performance. It will cover those small fine areas when using large areas at the latter. 
We explore the cues that the SAM predicted IoU has a similar value close to 1, without strong distinguish.

\begin{table}[]
\caption{Ablation Study on Individual Parts}
\resizebox{0.5\textwidth}{!}{
\begin{tabular}{ccc|ccc}
\hline
\textbf{Baseline} & \textbf{WSM} & \textbf{FCC} & \textbf{mIoU(\%)} & \textbf{mAcc(\%)} & \textbf{aAcc(\%)} \\ \hline
\checkmark                 &                   &                           & 45.10         & 57.44         & 83.53         \\
                    & \checkmark                 &                           & 44.40         & 54.48         & 82.27         \\
\checkmark                 & \checkmark                 &                           & 46.33         & 58.19         & 84.05         \\
\checkmark                 & \checkmark                 & \checkmark                         & 46.42         & 58.27         & 84.10         \\ \hline
\end{tabular}
}
\captionsetup{justification=centering}
\label{tab:ablation_all}
\end{table}

\begin{table}
\centering
\caption{Ablation Study on the Sorted Way of SAM Masks}
\begin{tabular}{c|ccc}
\hline
\textbf{Sorted Way}             & \textbf{mIoU (\%)} & \textbf{mAcc (\%)} & \textbf{aAcc (\%)} \\ \hline
IoU(\textit{large} $\rightarrow$ \textit{small})  & 45.70         & 57.63         & 83.70         \\
IoU(\textit{small} $\rightarrow$ \textit{large})  & 45.96         & 57.79         & 83.89         \\
Area(\textit{large} $\rightarrow$ \textit{small}) & 46.42         & 58.27         & 84.10         \\
Area(\textit{small} $\rightarrow$ \textit{large}) & 45.18         & 57.12         & 83.46         \\ \hline
\end{tabular}
\captionsetup{justification=centering}
\label{tab:ablation_sortway}
\end{table}

\begin{table}[]
\centering
\caption{Ablation Study on the Confused threshold of mask category on FoodSeg103.}
\begin{tabular}{c|ccc}
\hline
\textbf{Confused threshold}             & \textbf{mIoU (\%)} & \textbf{mAcc (\%)} & \textbf{aAcc (\%)} \\ \hline
0                        & 46.33      & 58.19       & 84.05                             \\
0.3                      & 46.33      & 58.19       & 84.05                             \\ 
0.5                      & 46.42      &  58.27      & 84.1                              \\
0.7                      & 46.25      & 58.18       & 84.05                             \\
0.9                      & 45.71      & 57.75       & 83.92                             \\ \hline
\end{tabular}
\captionsetup{justification=centering}
\label{tab:ablation_thres}
\end{table}

Before the merging process, if the number of the most category label is relatively low in a mask, we suppose the category of that mask is confused and filter them to the later merging process. 
Therefore, we also experiment with which threshold is effective for FoodSAM.
When the threshold $\tau$ equals zero, it means not using this practice.
The experiments on FoodSeg103 in shown in Tab.\ref{tab:ablation_thres} and UECFoodPix Complete shownn in \ref{tab:ablation_uec_thres}. 
Using the area of the original semantic mask to represent the confused mask can make a minor improvement in FoodSeg103, and a significant increase in UECFoodPix Complete. 
The reason behind this is that FoodSeg103 is the dataset for fine-grained ingredients, while UECFoodPix Complete only contains the food label, its label is coarse-grained. 
Therefore, when the number of confused labels is larger, our method achieves higher improvement.

\begin{table}
\centering
\caption{Ablation Study on the Confused threshold of mask category on Uecfoodpix Complete.}
\begin{tabular}{c|ccc}
\hline
\textbf{Confused threshold}             & \textbf{mIoU(\%)} & \textbf{mAcc(\%)} & \textbf{aAcc(\%)} \\ \hline
0                        & 65.61      & 77.56       & 88.20                             \\
0.3                      & 65.78      & 78.20       & 87.81                             \\ 
0.5                      & 65.81      &  78.21      & 87.85                              \\
0.7                      & 66.06      & 78.15       & 88.27                             \\
0.8                      & 66.14      & 78.01       & 88.47                             \\
0.9                      & 66.08      & 77.96       & 88.45                             \\ \hline
\end{tabular}
\captionsetup{justification=centering}
\label{tab:ablation_uec_thres}
\end{table}

\section{Conclusion}
This paper investigates the zero-shot capability of SAM for food image segmentation, a challenging task in the domain of food computing. 
The vanilla mask generation method of SAM alone falls short in capturing class-specific information, hindering accurate food item categorization. 
To address this limitation, we propose FoodSAM, a novel zero-shot framework that combines original semantic masks with SAM-generated category-agnostic masks to enhance semantic segmentation quality. 
Additionally, we leverage SAM's inherent instance-based masks to perform instance segmentation on food images. 
FoodSAM also incorporates object detection methodologies to detect non-food objects, allowing for panoptic segmentation. 
Furthermore, we extend our investigation to promptable segmentation, supporting various prompt variants. 
Our comprehensive evaluation on benchmark datasets demonstrates FoodSAM's state-of-the-art performance, 
affirming SAM's potential as an influential tool for food image segmentation. 
This work encompasses the first exploration of SAM in food segmentation, accomplishing instance, panoptic, and promotable segmentation on food images, and surpassing existing methods in performance.
 
\bibliographystyle{IEEEtran}
\bibliography{IEEEabrv, foodsam}

\begin{thebibliography}{10}
\providecommand{\url}[1]{#1}
\csname url@samestyle\endcsname
\providecommand{\newblock}{\relax}
\providecommand{\bibinfo}[2]{#2}
\providecommand{\BIBentrySTDinterwordspacing}{\spaceskip=0pt\relax}
\providecommand{\BIBentryALTinterwordstretchfactor}{4}
\providecommand{\BIBentryALTinterwordspacing}{\spaceskip=\fontdimen2\font plus
\BIBentryALTinterwordstretchfactor\fontdimen3\font minus
  \fontdimen4\font\relax}
\providecommand{\BIBforeignlanguage}[2]{{%
\expandafter\ifx\csname l@#1\endcsname\relax
\typeout{** WARNING: IEEEtran.bst: No hyphenation pattern has been}%
\typeout{** loaded for the language `#1'. Using the pattern for}%
\typeout{** the default language instead.}%
\else
\language=\csname l@#1\endcsname
\fi
#2}}
\providecommand{\BIBdecl}{\relax}
\BIBdecl

\bibitem{wolf2019huggingface}
T.~Wolf, L.~Debut, V.~Sanh, J.~Chaumond, C.~Delangue, A.~Moi, P.~Cistac,
  T.~Rault, R.~Louf, M.~Funtowicz \emph{et~al.}, ``Huggingface's transformers:
  State-of-the-art natural language processing,'' \emph{arXiv preprint
  arXiv:1910.03771}, 2019.

\bibitem{ruder2019transfer}
S.~Ruder, M.~E. Peters, S.~Swayamdipta, and T.~Wolf, ``Transfer learning in
  natural language processing,'' in \emph{Proceedings of the 2019 conference of
  the North American chapter of the association for computational linguistics:
  Tutorials}, 2019, pp. 15--18.

\bibitem{qiu2020pre}
X.~Qiu, T.~Sun, Y.~Xu, Y.~Shao, N.~Dai, and X.~Huang, ``Pre-trained models for
  natural language processing: A survey,'' \emph{Science China Technological
  Sciences}, vol.~63, no.~10, pp. 1872--1897, 2020.

\bibitem{kasneci2023chatgpt}
E.~Kasneci, K.~Se{\ss}ler, S.~K{\"u}chemann, M.~Bannert, D.~Dementieva,
  F.~Fischer, U.~Gasser, G.~Groh, S.~G{\"u}nnemann, E.~H{\"u}llermeier
  \emph{et~al.}, ``Chatgpt for good? on opportunities and challenges of large
  language models for education,'' \emph{Learning and Individual Differences},
  vol. 103, p. 102274, 2023.

\bibitem{wei2022emergent}
J.~Wei, Y.~Tay, R.~Bommasani, C.~Raffel, B.~Zoph, S.~Borgeaud, D.~Yogatama,
  M.~Bosma, D.~Zhou, D.~Metzler \emph{et~al.}, ``Emergent abilities of large
  language models,'' \emph{arXiv preprint arXiv:2206.07682}, 2022.

\bibitem{wei2022chain}
J.~Wei, X.~Wang, D.~Schuurmans, M.~Bosma, F.~Xia, E.~Chi, Q.~V. Le, D.~Zhou
  \emph{et~al.}, ``Chain-of-thought prompting elicits reasoning in large
  language models,'' \emph{Advances in Neural Information Processing Systems},
  vol.~35, pp. 24\,824--24\,837, 2022.

\bibitem{kirillov2023segment}
A.~Kirillov, E.~Mintun, N.~Ravi, H.~Mao, C.~Rolland, L.~Gustafson, T.~Xiao,
  S.~Whitehead, A.~C. Berg, W.-Y. Lo \emph{et~al.}, ``Segment anything,''
  \emph{arXiv preprint arXiv:2304.02643}, 2023.

\bibitem{10.1145/3600095}
\BIBentryALTinterwordspacing
J.~Song, Z.~Li, W.~Min, and S.~Jiang, ``Towards food image retrieval via
  generalization-oriented sampling and loss function design,'' \emph{ACM Trans.
  Multimedia Comput. Commun. Appl.}, may 2023, just Accepted. [Online].
  Available: \url{https://doi.org/10.1145/3600095}
\BIBentrySTDinterwordspacing

\bibitem{zhou2019application}
L.~Zhou, C.~Zhang, F.~Liu, Z.~Qiu, and Y.~He, ``Application of deep learning in
  food: a review,'' \emph{Comprehensive reviews in food science and food
  safety}, vol.~18, no.~6, pp. 1793--1811, 2019.

\bibitem{zhu2021deep}
L.~Zhu, P.~Spachos, E.~Pensini, and K.~N. Plataniotis, ``Deep learning and
  machine vision for food processing: A survey,'' \emph{Current Research in
  Food Science}, vol.~4, pp. 233--249, 2021.

\bibitem{wu2021large}
X.~Wu, X.~Fu, Y.~Liu, E.-P. Lim, S.~C. Hoi, and Q.~Sun, ``A large-scale
  benchmark for food image segmentation,'' in \emph{Proceedings of the 29th ACM
  International Conference on Multimedia}, 2021, pp. 506--515.

\bibitem{padilla2020survey}
R.~Padilla, S.~L. Netto, and E.~A. Da~Silva, ``A survey on performance metrics
  for object-detection algorithms,'' in \emph{2020 international conference on
  systems, signals and image processing (IWSSIP)}.\hskip 1em plus 0.5em minus
  0.4em\relax IEEE, 2020, pp. 237--242.

\bibitem{zou2023object}
Z.~Zou, K.~Chen, Z.~Shi, Y.~Guo, and J.~Ye, ``Object detection in 20 years: A
  survey,'' \emph{Proceedings of the IEEE}, 2023.

\bibitem{zhou2022simple}
X.~Zhou, V.~Koltun, and P.~Kr{\"a}henb{\"u}hl, ``Simple multi-dataset
  detection,'' in \emph{Proceedings of the IEEE/CVF Conference on Computer
  Vision and Pattern Recognition}, 2022, pp. 7571--7580.

\bibitem{okamoto2021uec}
K.~Okamoto and K.~Yanai, ``Uec-foodpix complete: A large-scale food image
  segmentation dataset,'' in \emph{Pattern Recognition. ICPR International
  Workshops and Challenges: Virtual Event, January 10--15, 2021, Proceedings,
  Part V}.\hskip 1em plus 0.5em minus 0.4em\relax Springer, 2021, pp. 647--659.

\bibitem{bubeck2023sparks}
S.~Bubeck, V.~Chandrasekaran, R.~Eldan, J.~Gehrke, E.~Horvitz, E.~Kamar,
  P.~Lee, Y.~T. Lee, Y.~Li, S.~Lundberg \emph{et~al.}, ``Sparks of artificial
  general intelligence: Early experiments with gpt-4,'' \emph{arXiv preprint
  arXiv:2303.12712}, 2023.

\bibitem{floridi2020gpt}
L.~Floridi and M.~Chiriatti, ``Gpt-3: Its nature, scope, limits, and
  consequences,'' \emph{Minds and Machines}, vol.~30, pp. 681--694, 2020.

\bibitem{brown2020language}
T.~Brown, B.~Mann, N.~Ryder, M.~Subbiah, J.~D. Kaplan, P.~Dhariwal,
  A.~Neelakantan, P.~Shyam, G.~Sastry, A.~Askell \emph{et~al.}, ``Language
  models are few-shot learners,'' \emph{Advances in neural information
  processing systems}, vol.~33, pp. 1877--1901, 2020.

\bibitem{hu2020gpt}
Z.~Hu, Y.~Dong, K.~Wang, K.-W. Chang, and Y.~Sun, ``Gpt-gnn: Generative
  pre-training of graph neural networks,'' in \emph{Proceedings of the 26th ACM
  SIGKDD International Conference on Knowledge Discovery \& Data Mining}, 2020,
  pp. 1857--1867.

\bibitem{radford2021learning}
A.~Radford, J.~W. Kim, C.~Hallacy, A.~Ramesh, G.~Goh, S.~Agarwal, G.~Sastry,
  A.~Askell, P.~Mishkin, J.~Clark \emph{et~al.}, ``Learning transferable visual
  models from natural language supervision,'' in \emph{International conference
  on machine learning}.\hskip 1em plus 0.5em minus 0.4em\relax PMLR, 2021, pp.
  8748--8763.

\bibitem{li2021supervision}
Y.~Li, F.~Liang, L.~Zhao, Y.~Cui, W.~Ouyang, J.~Shao, F.~Yu, and J.~Yan,
  ``Supervision exists everywhere: A data efficient contrastive language-image
  pre-training paradigm,'' \emph{arXiv preprint arXiv:2110.05208}, 2021.

\bibitem{bommasani2021opportunities}
R.~Bommasani, D.~A. Hudson, E.~Adeli, R.~Altman, S.~Arora, S.~von Arx, M.~S.
  Bernstein, J.~Bohg, A.~Bosselut, E.~Brunskill \emph{et~al.}, ``On the
  opportunities and risks of foundation models,'' \emph{arXiv preprint
  arXiv:2108.07258}, 2021.

\bibitem{yuan2021florence}
L.~Yuan, D.~Chen, Y.-L. Chen, N.~Codella, X.~Dai, J.~Gao, H.~Hu, X.~Huang,
  B.~Li, C.~Li \emph{et~al.}, ``Florence: A new foundation model for computer
  vision,'' \emph{arXiv preprint arXiv:2111.11432}, 2021.

\bibitem{touvron2023llama}
H.~Touvron, T.~Lavril, G.~Izacard, X.~Martinet, M.-A. Lachaux, T.~Lacroix,
  B.~Rozi{\`e}re, N.~Goyal, E.~Hambro, F.~Azhar \emph{et~al.}, ``Llama: Open
  and efficient foundation language models,'' \emph{arXiv preprint
  arXiv:2302.13971}, 2023.

\bibitem{pal1993review}
N.~R. Pal and S.~K. Pal, ``A review on image segmentation techniques,''
  \emph{Pattern recognition}, vol.~26, no.~9, pp. 1277--1294, 1993.

\bibitem{cheng2001color}
H.-D. Cheng, X.~H. Jiang, Y.~Sun, and J.~Wang, ``Color image segmentation:
  advances and prospects,'' \emph{Pattern recognition}, vol.~34, no.~12, pp.
  2259--2281, 2001.

\bibitem{mcguinness2010comparative}
K.~McGuinness and N.~E. O’connor, ``A comparative evaluation of interactive
  segmentation algorithms,'' \emph{Pattern Recognition}, vol.~43, no.~2, pp.
  434--444, 2010.

\bibitem{mortensen1998interactive}
E.~N. Mortensen and W.~A. Barrett, ``Interactive segmentation with intelligent
  scissors,'' \emph{Graphical models and image processing}, vol.~60, no.~5, pp.
  349--384, 1998.

\bibitem{liu2011entropy}
M.-Y. Liu, O.~Tuzel, S.~Ramalingam, and R.~Chellappa, ``Entropy rate superpixel
  segmentation,'' in \emph{CVPR 2011}.\hskip 1em plus 0.5em minus 0.4em\relax
  IEEE, 2011, pp. 2097--2104.

\bibitem{haralick1985image}
R.~M. Haralick and L.~G. Shapiro, ``Image segmentation techniques,''
  \emph{Computer vision, graphics, and image processing}, vol.~29, no.~1, pp.
  100--132, 1985.

\bibitem{minaee2021image}
S.~Minaee, Y.~Boykov, F.~Porikli, A.~Plaza, N.~Kehtarnavaz, and D.~Terzopoulos,
  ``Image segmentation using deep learning: A survey,'' \emph{IEEE transactions
  on pattern analysis and machine intelligence}, vol.~44, no.~7, pp.
  3523--3542, 2021.

\bibitem{guo2018review}
Y.~Guo, Y.~Liu, T.~Georgiou, and M.~S. Lew, ``A review of semantic segmentation
  using deep neural networks,'' \emph{International journal of multimedia
  information retrieval}, vol.~7, pp. 87--93, 2018.

\bibitem{wang2018understanding}
P.~Wang, P.~Chen, Y.~Yuan, D.~Liu, Z.~Huang, X.~Hou, and G.~Cottrell,
  ``Understanding convolution for semantic segmentation,'' in \emph{2018 IEEE
  winter conference on applications of computer vision (WACV)}.\hskip 1em plus
  0.5em minus 0.4em\relax Ieee, 2018, pp. 1451--1460.

\bibitem{hao2020brief}
S.~Hao, Y.~Zhou, and Y.~Guo, ``A brief survey on semantic segmentation with
  deep learning,'' \emph{Neurocomputing}, vol. 406, pp. 302--321, 2020.

\bibitem{fu2019dual}
J.~Fu, J.~Liu, H.~Tian, Y.~Li, Y.~Bao, Z.~Fang, and H.~Lu, ``Dual attention
  network for scene segmentation,'' in \emph{Proceedings of the IEEE/CVF
  conference on computer vision and pattern recognition}, 2019, pp. 3146--3154.

\bibitem{yuan2018ocnet}
Y.~Yuan, L.~Huang, J.~Guo, C.~Zhang, X.~Chen, and J.~Wang, ``Ocnet: Object
  context network for scene parsing,'' \emph{arXiv preprint arXiv:1809.00916},
  2018.

\bibitem{zheng2021rethinking}
S.~Zheng, J.~Lu, H.~Zhao, X.~Zhu, Z.~Luo, Y.~Wang, Y.~Fu, J.~Feng, T.~Xiang,
  P.~H. Torr \emph{et~al.}, ``Rethinking semantic segmentation from a
  sequence-to-sequence perspective with transformers,'' in \emph{Proceedings of
  the IEEE/CVF conference on computer vision and pattern recognition}, 2021,
  pp. 6881--6890.

\bibitem{hafiz2020survey}
A.~M. Hafiz and G.~M. Bhat, ``A survey on instance segmentation: state of the
  art,'' \emph{International journal of multimedia information retrieval},
  vol.~9, no.~3, pp. 171--189, 2020.

\bibitem{he2017mask}
K.~He, G.~Gkioxari, P.~Doll{\'a}r, and R.~Girshick, ``Mask r-cnn,'' in
  \emph{Proceedings of the IEEE international conference on computer vision},
  2017, pp. 2961--2969.

\bibitem{bolya2019yolact}
D.~Bolya, C.~Zhou, F.~Xiao, and Y.~J. Lee, ``Yolact: Real-time instance
  segmentation,'' in \emph{Proceedings of the IEEE/CVF international conference
  on computer vision}, 2019, pp. 9157--9166.

\bibitem{kirillov2019panoptic}
A.~Kirillov, K.~He, R.~Girshick, C.~Rother, and P.~Doll{\'a}r, ``Panoptic
  segmentation,'' in \emph{Proceedings of the IEEE/CVF conference on computer
  vision and pattern recognition}, 2019, pp. 9404--9413.

\bibitem{xiong2019upsnet}
Y.~Xiong, R.~Liao, H.~Zhao, R.~Hu, M.~Bai, E.~Yumer, and R.~Urtasun, ``Upsnet:
  A unified panoptic segmentation network,'' in \emph{Proceedings of the
  IEEE/CVF Conference on Computer Vision and Pattern Recognition}, 2019, pp.
  8818--8826.

\bibitem{elharrouss2021panoptic}
O.~Elharrouss, S.~Al-Maadeed, N.~Subramanian, N.~Ottakath, N.~Almaadeed, and
  Y.~Himeur, ``Panoptic segmentation: A review,'' \emph{arXiv preprint
  arXiv:2111.10250}, 2021.

\bibitem{liu2023pre}
P.~Liu, W.~Yuan, J.~Fu, Z.~Jiang, H.~Hayashi, and G.~Neubig, ``Pre-train,
  prompt, and predict: A systematic survey of prompting methods in natural
  language processing,'' \emph{ACM Computing Surveys}, vol.~55, no.~9, pp.
  1--35, 2023.

\bibitem{luddecke2022image}
T.~L{\"u}ddecke and A.~Ecker, ``Image segmentation using text and image
  prompts,'' in \emph{Proceedings of the IEEE/CVF Conference on Computer Vision
  and Pattern Recognition}, 2022, pp. 7086--7096.

\bibitem{liu2021gpt}
X.~Liu, Y.~Zheng, Z.~Du, M.~Ding, Y.~Qian, Z.~Yang, and J.~Tang, ``Gpt
  understands, too,'' \emph{arXiv preprint arXiv:2103.10385}, 2021.

\bibitem{lu2021pretrained}
K.~Lu, A.~Grover, P.~Abbeel, and I.~Mordatch, ``Pretrained transformers as
  universal computation engines,'' \emph{arXiv preprint arXiv:2103.05247},
  vol.~1, 2021.

\bibitem{MIN2022100484}
\BIBentryALTinterwordspacing
W.~Min, C.~Liu, L.~Xu, and S.~Jiang, ``Applications of knowledge graphs for
  food science and industry,'' \emph{Patterns}, vol.~3, no.~5, p. 100484, 2022.
  [Online]. Available:
  \url{https://www.sciencedirect.com/science/article/pii/S2666389922000691}
\BIBentrySTDinterwordspacing

\bibitem{min2023large}
W.~Min, Z.~Wang, Y.~Liu, M.~Luo, L.~Kang, X.~Wei, X.~Wei, and S.~Jiang, ``Large
  scale visual food recognition,'' \emph{IEEE Transactions on Pattern Analysis
  and Machine Intelligence}, 2023.

\bibitem{Min-ISIA-500-MM2020}
W.~Min, L.~Liu, Z.~Wang, Z.~Luo, X.~Wei, X.~Wei, and S.~Jiang, ``Isia food-500:
  A dataset for large-scale food recognition via stacked global-local attention
  network,'' in \emph{Proceedings of the 28th ACM International Conference on
  Multimedia}, 2020.

\bibitem{klotz2021fine}
J.~Klotz, V.~Rengarajan, and A.~C. Sankaranarayanan, ``Fine-grain prediction of
  strawberry freshness using subsurface scattering,'' in \emph{Proceedings of
  the IEEE/CVF International Conference on Computer Vision}, 2021, pp.
  2328--2336.

\bibitem{huang2019ccnet}
Z.~Huang, X.~Wang, L.~Huang, C.~Huang, Y.~Wei, and W.~Liu, ``Ccnet: Criss-cross
  attention for semantic segmentation,'' in \emph{Proceedings of the IEEE/CVF
  international conference on computer vision}, 2019, pp. 603--612.

\bibitem{wang2022swin}
Q.~Wang, X.~Dong, R.~Wang, and H.~Sun, ``Swin transformer based pyramid pooling
  network for food segmentation,'' in \emph{2022 IEEE 2nd International
  Conference on Software Engineering and Artificial Intelligence (SEAI)}.\hskip
  1em plus 0.5em minus 0.4em\relax IEEE, 2022, pp. 64--68.

\bibitem{honbu2022unseen}
Y.~Honbu and K.~Yanai, ``Unseen food segmentation,'' in \emph{Proceedings of
  the 2022 International Conference on Multimedia Retrieval}, 2022, pp. 19--23.

\bibitem{sinha2023transferring}
G.~Sinha, K.~Parmar, H.~Azimi, A.~Tai, Y.~Chen, A.~Wong, and P.~Xi,
  ``Transferring knowledge for food image segmentation using transformers and
  convolutions,'' \emph{arXiv preprint arXiv:2306.09203}, 2023.

\bibitem{dosovitskiy2020image}
A.~Dosovitskiy, L.~Beyer, A.~Kolesnikov, D.~Weissenborn, X.~Zhai,
  T.~Unterthiner, M.~Dehghani, M.~Minderer, G.~Heigold, S.~Gelly \emph{et~al.},
  ``An image is worth 16x16 words: Transformers for image recognition at
  scale,'' \emph{arXiv preprint arXiv:2010.11929}, 2020.

\bibitem{he2022masked}
K.~He, X.~Chen, S.~Xie, Y.~Li, P.~Doll{\'a}r, and R.~Girshick, ``Masked
  autoencoders are scalable vision learners,'' in \emph{Proceedings of the
  IEEE/CVF conference on computer vision and pattern recognition}, 2022, pp.
  16\,000--16\,009.

\bibitem{vaswani2017attention}
A.~Vaswani, N.~Shazeer, N.~Parmar, J.~Uszkoreit, L.~Jones, A.~N. Gomez,
  {\L}.~Kaiser, and I.~Polosukhin, ``Attention is all you need,''
  \emph{Advances in neural information processing systems}, vol.~30, 2017.

\bibitem{zhang2023recognize}
Y.~Zhang, X.~Huang, J.~Ma, Z.~Li, Z.~Luo, Y.~Xie, Y.~Qin, T.~Luo, Y.~Li, S.~Liu
  \emph{et~al.}, ``Recognize anything: A strong image tagging model,''
  \emph{arXiv preprint arXiv:2306.03514}, 2023.

\bibitem{zou2023segment}
X.~Zou, J.~Yang, H.~Zhang, F.~Li, L.~Li, J.~Wang, L.~Wang, J.~Gao, and Y.~J.
  Lee, ``Segment everything everywhere all at once,'' 2023.

\bibitem{chen2023semantic}
J.~Chen, Z.~Yang, and L.~Zhang, ``Semantic segment anything,''
  \url{https://github.com/fudan-zvg/Semantic-Segment-Anything}, 2023.

\bibitem{zhao2023fast}
X.~Zhao, W.~Ding, Y.~An, Y.~Du, T.~Yu, M.~Li, M.~Tang, and J.~Wang, ``Fast
  segment anything,'' \emph{arXiv preprint arXiv:2306.12156}, 2023.

\bibitem{rother2004grabcut}
C.~Rother, V.~Kolmogorov, and A.~Blake, ``" grabcut" interactive foreground
  extraction using iterated graph cuts,'' \emph{ACM transactions on graphics
  (TOG)}, vol.~23, no.~3, pp. 309--314, 2004.

\bibitem{ege2019new}
T.~Ege, W.~Shimoda, and K.~Yanai, ``A new large-scale food image segmentation
  dataset and its application to food calorie estimation based on grains of
  rice,'' in \emph{Proceedings of the 5th international workshop on multimedia
  assisted dietary management}, 2019, pp. 82--87.

\bibitem{chen2018encoder}
L.-C. Chen, Y.~Zhu, G.~Papandreou, F.~Schroff, and H.~Adam, ``Encoder-decoder
  with atrous separable convolution for semantic image segmentation,'' in
  \emph{Proceedings of the European conference on computer vision (ECCV)},
  2018, pp. 801--818.

\bibitem{dong2021windows}
X.~Dong, W.~Wang, H.~Li, and Q.~Cai, ``Windows attention based pyramid network
  for food segmentation,'' in \emph{2021 IEEE 7th International Conference on
  Cloud Computing and Intelligent Systems (CCIS)}.\hskip 1em plus 0.5em minus
  0.4em\relax IEEE, 2021, pp. 213--217.

\bibitem{xiao2018unified}
T.~Xiao, Y.~Liu, B.~Zhou, Y.~Jiang, and J.~Sun, ``Unified perceptual parsing
  for scene understanding,'' in \emph{Proceedings of the European conference on
  computer vision (ECCV)}, 2018, pp. 418--434.

\bibitem{battini2023segmented}
E.~Battini~S{\"o}nmez, S.~Memi{\c{s}}, B.~Arslan, and O.~Z. Batur, ``The
  segmented uec food-100 dataset with benchmark experiment on food detection,''
  \emph{Multimedia Systems}, pp. 1--9, 2023.

\bibitem{sharma2021gourmetnet}
U.~Sharma, B.~Artacho, and A.~Savakis, ``Gourmetnet: Food segmentation using
  multi-scale waterfall features with spatial and channel attention,''
  \emph{Sensors}, vol.~21, no.~22, p. 7504, 2021.

\bibitem{aguilar2022bayesian}
E.~Aguilar, B.~Nagarajan, B.~Remeseiro, and P.~Radeva, ``Bayesian deep learning
  for semantic segmentation of food images,'' \emph{Computers and Electrical
  Engineering}, vol. 103, p. 108380, 2022.

\end{thebibliography}

\end{document}